\pdfoutput=1
\PassOptionsToPackage{dvipsnames,svgnames,table}{xcolor} 

\documentclass[11pt]{article}

\usepackage[preprint]{acl}

\usepackage{times}
\usepackage{latexsym}

\usepackage[T1]{fontenc}

\usepackage[utf8]{inputenc}

\usepackage{microtype}

\usepackage{inconsolata}

\usepackage{color,soul}
\usepackage{booktabs}
\usepackage{multirow}
\usepackage{float}
\usepackage{enumitem}

\usepackage{graphicx}

\usepackage{tcolorbox}

\usepackage{listings}
\title{Multi-domain Multilingual Sentiment Analysis in Industry: Predicting Aspect-based Opinion Quadruples}

\author{\textbf{Benjamin White} \hspace{1cm}
\textbf{Anastasia Shimorina} \\
\\
Orange Innovation, Lannion, France
\\
\\
\texttt{firstname.lastname@orange.com}
}

\begin{document}
\maketitle
\begin{abstract}
This paper explores the design of an aspect-based sentiment analysis system using large language models (LLMs) for real-world use. We focus on quadruple opinion extraction---identifying aspect categories, sentiment polarity, targets, and opinion expressions from text data across different domains and languages. We investigate whether a single fine-tuned model can effectively handle multiple domain-specific taxonomies simultaneously. We demonstrate that a combined multi-domain model achieves performance comparable to specialized single-domain models while reducing operational complexity. We also share lessons learned for handling non-extractive predictions and evaluating various failure modes when developing LLM-based systems for structured prediction tasks.
\end{abstract}

\section{Introduction}

\begin{table*}
\small
\renewcommand{\arraystretch}{1.5} 
    \centering
    \begin{tabular}{lcccc}
    \toprule
         \textbf{Text} & \textbf{Target} & \textbf{Aspect Category} & \textbf{Sentiment} & \textbf{Domain} \\
    \midrule
        I use the service for 1 year, and \hl{the \textbf{cost}? Don't even ask me} & cost & Price & Negative & CR \\
        \hl{frequent mandatory software updates} on my phone & NULL & Usability & Negative & PS \\
        ... \hl{from one \textbf{boss} to the next we can go from good to bad}... & boss & Management & Negative & HR \\
        \multirow{2}{23em}{Personally, \hl{my \textbf{colleagues} seem qualified to me} and \hl{I enjoy working with my \textbf{team}}. Hope that will last.} & colleagues & Job Skills & Positive & HR \\
         & team & General & Positive & HR \\
    \bottomrule
    \end{tabular}
    \caption{Excerpts from user reviews and employee surveys. Opinion expressions are highlighted in yellow; targets are in bold. The implicit target (NULL) in the second example refers to a mobile application. Domains: Customer Relations (CR), Products and Services (PS), Human Resources (HR).}
    \label{tab:OpinionTool_examples}
\end{table*}

\begin{table*}
\small
\centering
\begin{tabular}{lrp{7cm}p{2.6cm}p{2.2cm}}
\toprule
\textbf{Dataset} & \textbf{Asp. cat.} & \textbf{Description} & \textbf{Languages} & \textbf{Style} \\
\midrule
PS & 14 & Customer feedback about our company's products and services, obtained from surveys, web forms, chatbots & en, fr, ar, es, ro, pl, nl & concise, informal \\
HR & 17 & Internal employee surveys concerning all HR related topics & en, fr & detailed, formal \\
CR & 17 & Customer feedback about their experience with our company's customer support teams and advisors & fr & detailed, informal \\
\bottomrule
\end{tabular}
\caption{Overview of the internal datasets used in experiments. Asp. cat.: the number of aspect categories in the domain taxonomy. In total, all three domains have 48 aspect categories.}
\label{tab:datasetstable}
\end{table*}

In today's data-driven landscape, understanding stakeholder sentiment is crucial for organisational success. Our company processes hundreds of thousands of user reviews and employee surveys annually across different countries, containing valuable insights about product features, workplace conditions, and company policies. We have an internal solution, OpinionTool, that analyses textual data and performs opinion mining. OpinionTool makes use of aspect-based sentiment analysis \cite[ABSA]{huliu2004, pontiki-etal-2014-semeval} that offers a granular approach by identifying and analysing sentiment towards distinct aspects within text data. In ABSA, extracting meaningful insights from multi-sentence texts is a complex task, especially when dealing with implicit targets, diverse syntactic structures, and varying languages. To address these challenges, we propose quadruple opinion extraction, which identifies aspect category, sentiment polarity, target, and opinion expression.

Notably, we introduce a novel definition of opinion expression, which refers to a sequence of words that justifies sentiment polarity and may include the target itself. Given the text ``\textit{I think that from one boss to the next we can go from good to bad. It all depends on their management style and communication.}'', the aim of ABSA is to identify four elements: the specific target of the opinion (\textit{boss}), the broader aspect category associated with the target (\textit{Management}), the opinion expression (\textit{from one boss to the next we can go from good to bad}), and the sentiment polarity linked to the opinion expression (\textit{negative}). Once opinion quadruples are identified in OpinionTool, they are used by different professionals in the company for product development and for Human Resources (HR) policy improvements.

While OpinionTool\ has been successfully used in production for several years, a key challenge lies in the aspect category detection task, which relies on classification taxonomies defined by end users. For instance, in the HR domain, the taxonomy includes 17 aspect categories such as Management, Job Skills, Working Conditions, Remote Work, etc. Currently, OpinionTool\ supports three distinct taxonomies across different domains: Products and Services (PS), HR, and Customer Relations (CR); see Table~\ref{tab:datasetstable}. However, the dynamic nature of user reviews, evolving products, and changing working conditions requires frequent updates to these taxonomies. This results in the need to develop and maintain separate models for each taxonomy or use case, which is both resource-intensive and operationally complex. Additionally, some users require overlapping or customized subsets of the existing taxonomies, further complicating the modeling process. Furthermore, vulnerabilities discovered in third-party libraries used by models necessitate retraining all three models with the updated, secure library version. To overcome these challenges and minimise retraining overhead, we propose the development of a generic, multi-domain model capable of handling multiple taxonomies simultaneously.

In this study, we explore fine-tuning LLMs for the ABSA quadruple prediction task on different domains and languages given domain-specific and combined taxonomies for aspect categories.

\section{Background and Related Work}
\paragraph{ABSA Subtasks} 
ABSA encompasses various subtasks, depending on the combinations of elements predicted from the quadruple \{\textit{aspect category}, \textit{sentiment}, \textit{target}, \textit{opinion expression}\}: sentiment classification, target aspect detection, aspect sentiment triplet extraction \cite{peng2020knowing, li2023aspectsentiment}, and others \cite{chebolu-etal-2023-review}. Our approach focuses on a subtask closely related to the aspect-sentiment quadruple prediction \cite{cai-etal-2021-aspect, zhang-etal-2021-aspect-sentiment}, where the goal is to extract all four elements, such as \{target: \textit{pizza}, opinion term/expression: \textit{delicious}, aspect category: \textit{Food}, sentiment: \textit{positive}\} from the sentence ``\textit{The pizza is delicious.}'' In this work, we introduce a novel ABSA-quad task by including the target itself in the opinion expression, which is particularly useful when analyzing multi-sentence long texts. This approach also accommodates diverse syntactic and morphological structures across languages and facilitates the opinion expression extraction from sentences where the target and the opinion expression forms the same syntactic group (e.g., ``\textit{I enjoy working with my team}''). More examples can be seen in Table~\ref{tab:OpinionTool_examples}. Extracting the entire opinion expression together with the target is more advantageous for visual analysis by end-users and is more useful for relevant downstream applications such as opinion mining, opinion expression clustering and summarization, and in-depth opinion analysis.

\paragraph{ABSA Modeling}
Earlier work mainly treated ABSA as a classification task with predominantly BERT-based approaches. When the quadruple extraction task was first introduced, two main directions were proposed. The first approach, proposed by \cite{zhang-etal-2021-aspect-sentiment}, employs a generation-based model that enables end-to-end extraction of quadruples through sentence paraphrasing techniques. This generative approach particularly emphasizes the order-free nature of quadruples. The generation-based framework has gained significant attention, with subsequent studies by \citet{hu-etal-2022-improving-aspect, peper-wang-2022-generative}.
The second approach by \citet{cai-etal-2021-aspect} implements a two-stage methodology where aspect and opinion terms are initially extracted, followed by the classification of aspect categories and sentiment polarity.

Nowadays generative methods have become mainstream for ABSA tasks \cite{wang-etal-2023-generative, zhang-etal-2024-self-training, bai-etal-2024-bvsp, zhu-etal-2024-pinpointing}.
\citet{wang-etal-2023-generative} made use of data augmentation for quadruple prediction; \citet{zhu-etal-2024-pinpointing} combined a diffusion component with supervised contrastive learning; \citet{su-etal-2025-unified} employed graph diffusion convolution networks using syntactic and semantic features.

Recently \citet{bai-etal-2024-compound} have explored prompting with LLMs (ChatGPT and Llama-3 \cite{grattafiori2024llama3herdmodels}) for the quadruple task.

\paragraph{Multi-task and Multi-Domain ABSA}
Several unified generative approaches were proposed for handling different ABSA subtasks at once based on multi-task prompt tuning \cite{gao-etal-2022-lego, wang2024unified}.

\citet{yang-etal-2024-faima} built a framework for aspect-sentiment extraction across nine different domains. \citet{scaria-etal-2024-instructabsa} proposed both multi-task and multi-domain instruction tuning for ABSA tasks including quadruple opinion extraction. 
\citet{fei-etal-2023-reasoning} provided a reasoning-based solution with chain-of-thought prompting for quadruple extraction also applicable to the multi-domain setting.
However, empirical studies targeting multi-domain quadruple ABSA extraction remain largely limited to English \cite{cai2025memd} and well-known domains of Laptop and Restaurant from the SemEval challenges \cite{pontiki-etal-2014-semeval}.

\section{Experiments}
\label{sec:experiments}
\subsection{Datasets}

As described above, the existing OpinionTool\ workflow for ABSA involves annotating large volumes of textual data from various sources such as client feedback or employee satisfaction surveys. Consequently we have an internal collection of several large datasets from across our company's operations. Each dataset has been annotated with a similar \textit{sentiment taxonomy} (\texttt{positive, negative, neutral, mixed}) but with a variable \textit{domain-specific aspect category taxonomy}. An individual text may have any number of annotated quadruples.

All annotations were produced by human annotators with domain expertise, who were instructed to follow a pre-established annotation guidebook.

For our experiments we curated 3 different datasets from across our company's operations to obtain a range of business domains, languages, and complexity (see Table~\ref{tab:datasetstable}). We include a more detailed statistical description of the implicit and explicit opinion targets and the distribution of the number of quadruples across our datasets in Appendix~\ref{subsec:datasetstats}. 

During data preparation we noticed that spelling mistakes or informal writing styles were common in the 3 datasets. While this does not typically present a problem for encoder-based modeling approaches, which tend to classify using token-level embeddings for example, we anticipated that generative LLM approaches may attempt to ``correct'' their outputs---we discuss our strategies for evaluating this model behavior in Section \ref{subsec:evalmetrics}.

Public multilingual benchmarks for quad-ABSA are scarce, with no datasets available for multilingual quadruple extraction or our novel definition of opinion expression—both of which are key contributions of our work. The closest existing dataset is M-ABSA \citep{wu2025mabsa}, a multilingual dataset created by translating data and focused on triple extraction (aspect category, target, and sentiment). Although it does not fully align with our focus on quadruple extraction and complex opinion expressions, we used it to evaluate the feasibility of our approach on public multilingual data. The details of our experiments and results are provided in Appendix~\ref{app:m-absa-experiments}.

\begin{table*}[ht!]
\small
\centering
\begin{tabular}{ll}
\toprule
\textbf{Type} & \textbf{Description} \\
\midrule
Invalid JSON & The model output string does not load to a valid JSON object \\
Incorrect keys & Valid JSON object but some keys do not correspond to desired schema \\
Invalid sentiment label & Valid JSON object but the sentiment label is not in the desired extraction schema \\
Invalid aspect category label & Valid JSON object but the aspect category label is not in the desired extraction schema \\
Non-extractive target & Valid JSON but predicted target is not a direct substring of input text \\
Non-extractive opinion expression & Valid JSON but predicted opinion expression is not a direct substring of input text \\
\bottomrule
\end{tabular}
\caption{Possible ``failure modes'' when performing structured extractive ABSA with a generative model.}
\label{tab:failuremodes}
\end{table*}

\subsection{Task formulation and data preparation}

To model the ABSA task with generative LLMs, we formulate the model training as a supervised fine-tuning (SFT) task on instruction-response pairs. An example of our prompts used for training is in Appendix \ref{subsec:sampleprompt}, which we summarize briefly here.

To prepare the SFT instructions we first define a generic ABSA user prompt template which describes the vocabulary and structure of the task, and which is reused across all 3 dataset domains. The prompt contains empty variables for the \textit{domain-specific aspect categories} and for a \textit{domain-specific one-shot example}.

Then, for each of the different datasets, we create the following domain-specific variables: \textit{(i)} a system prompt; \textit{(ii)} a list of the aspect category labels to use during classification, along with a brief description of each label; \textit{(iii)} a representative one-shot example. For each dataset we format our generic prompt with these domain-specific variables, and then insert the input text to be analysed by the model.

We experimented with three settings for language selection in instruction prompts: (1) using only English prompts, (2) using only French prompts (as French is the dominant language in the dataset in terms of sample count), and (3) using a mix of English and French prompts based on the language of the data sample (French prompts for French samples, and English prompts for English and other language samples).

\subsection{Single domain vs. multi-domain systems}
\label{subsec:indivvscombineddomains}

A key question we wanted to address is whether a small LLM could be trained to perform in real-world use as a \textbf{multi-domain generalized extraction model}, or whether it would be necessary to deploy a specific fine-tuned model for each single domain.
To answer this question, for each pre-trained model we systematically fine-tuned 4 different models by adjusting the training data used: we created 3 \textbf{single domain models} by fine-tuning on each of the 3 different domain datasets, and then also created a 4th \textbf{multi-domain model} by fine-tuning on all 3 datasets combined.

For each dataset, we subsequently evaluated this multi-domain model against the corresponding single domain model.

\subsection{Model selection and training}

To align with technical constraints given to us by project stakeholders, we focused our experiments on small models that could be deployed locally with minimal hardware requirements. We selected the recent Qwen2.5 family of models \cite{qwen2025qwen25technicalreport} for its multilingual pre-training that covers the languages appearing in our company's datasets, and its extensive post-training using structured data. We performed all experiments using the 0.5B, 1.5B, and 3B parameter Instruct models. Training details are given in Appendix~\ref{app:finetuninghyperparams}.

Based on the envisioned business use case for our trained models---frequent requests involving large volumes of data---we selected GPT-4o mini (2024-07-18) as a realistic cost-effective API model for an additional baseline in our studies. All requests were made using OpenAI's structured generation API\footnote{\texttt{openai==1.55.0}; the Pydantic schema is in App. \ref{subsec:gptdataschema}.}.

\subsection{Evaluation and metrics}
\label{subsec:evalmetrics}

\begin{table*}[ht!]
\centering
\small
\begin{tabular}{@{}lrrrrrrrr@{}}
& & & \multicolumn{5}{c}{Failure modes} \\
\cline{3-8}
& & JSON & Keys & Sent & AspCat & NETarg & NEOpExp & Total preds \\
\hline
GPT-4o mini & NA & 0 & 0 & 0 & 0 & 202 & 300 & 5,555 \\
\hline
Base & 0.5B & 3 & 0 & 0 & 276 & 3,716 & 4,602 & 4,936 \\
& 1.5B & 12 & 1 & 0 & 21 & 1,793 & 4,000 & 5,129 \\
& 3B & 2 & 1 & 1 & 18 & 597 & 734 & 4,152 \\
\hline
Single domain & 0.5B & 2 & 0 & 0 & 0 & 104 & 332 & 5,816 \\
& 1.5B & 1 & 0 & 0 & 0 & 63 & 238 & 5,764 \\
& 3B & 0 & 0 & 0 & 2 & 42 & 146 & 5,909 \\
\hline
Multi-domain & 0.5B & 6 & 0 & 0 & 0 & 114 & 288 & 5,289 \\
& 1.5B & 2 & 0 & 0 & 0 & 51 & 197 & 5,662 \\
& 3B & 2 & 0 & 0 & 0 & 29 & 134 & 5,911 \\
\bottomrule
\end{tabular}
\caption{Failure modes for models during output post-processing (all models are Qwen2.5 apart from the GPT-4o mini row). The results shown are from inference on the \textbf{PS domain} dataset---the results for all other domains show similar patterns. For this PS domain test dataset there are 5,796 total reference quadruples, we show the total number of predictions made by each model in the \textit{Total preds} column. All other failure mode abbreviations refer to the Types in Table~\ref{tab:failuremodes}. The parameter size of GPT-4o mini is not publicly available but estimated to be more than 3B.}
\label{tab:resultsfailuremode}
\end{table*}

We evaluated two complementary dimensions of our models: the technical performance, namely the ability to actually generate valid parsable outputs that respect the various structural constraints, and the task performance, namely how well the predictions match the reference quadruples.

\paragraph{Technical performance} Deploying generative LLM-based solutions in production presents several difficulties linked to the open-ended nature of the generation process. For a given input text, we expect our trained model to output a generated text string that corresponds to a valid JSON object. Then, for downstream processing, we expect schema keys and classification labels to be reliable. Finally, to avoid hallucinations and to be able to present model predictions to the end user, we require model predictions to be purely extractive, i.e. correspond to substrings that actually appear in the input text.

To gain a better insight into the behavior of our proposed approach, we conceptualize and evaluate these requirements in terms of several possible ``failure modes'' as described in Table~\ref{tab:failuremodes}.

All raw model predictions were processed with the \texttt{json-repair} Python library \cite{baccianella2025jsonrepair}; in all results hereafter, \textit{invalid JSON} thus refers to any raw model output that was not able to be parsed by this tool.

\paragraph{Task performance}
To align with the existing ABSA research literature, we consider a model prediction as true only if it matches on \textbf{all 4} of the quadruple values. In what follows we refer to this evaluation criterion as the \textit{strict evaluation}.

We also consider a second evaluation criterion, which we refer to as \textit{relaxed evaluation}, where we consider a model prediction as true if it matches \textbf{strictly} on 3 values of the reference quadruple (target, sentiment, and aspect category, as before), but where for the opinion expression extraction we allow a \textbf{partial overlap} with the reference.

\section{Results}
\label{sec:results}

\subsection{Technical performance}

To measure the capabilities of small LLMs for our complex structured extraction task, we first present a breakdown of the various failure modes identified in Table~\ref{tab:failuremodes}. For each model and each dataset being tested, we record the various errors encountered during all stages of our post-processing. Table~\ref{tab:resultsfailuremode} shows a representative summary for the PS dataset (here, and in all other results, \textit{Base} refers to models without any fine-tuning).

Examining the performance of GPT-4o mini, while the structured generation API ensures 0 errors in the JSON schema related columns, we note that---despite the comparatively much larger model size---we obtain 202 and 300 instances of non-extractive behavior (\textit{NETarg} and \textit{NEOpExp} columns respectively).

Remarkably, with the post-processing described in Section \ref{subsec:evalmetrics}, we find that all models in the Qwen2.5 family can reliably be used for structured JSON outputs and following schemas specified via prompting. In general, while all models respect the imposed list of sentiment labels during classification, we note that the aspect category classification task generates 276 unknown labels in the case of the smallest base model. Importantly, fine-tuning consistently improves this performance across all model sizes, suggesting that even the 0.5B models can be trained towards structured prediction.

Turning to the 2 extractive tasks---predicting a target and an opinion expression that must be substrings within the input text---we note firstly that performance improves significantly with increasing model size. While base models without fine-tuning invariably try to correct the various spelling mistakes or domain-specific terminology encountered in the datasets, our fine-tuning effectively teaches all models in a size-dependent manner to extract without modification. Interestingly, even the smallest 0.5B fine-tuned model achieves performance comparable to GPT-4o mini on these 2 subtasks, while the 3B models make very few such unwanted corrections overall.

Crucially for our study, the results obtained for the multi-domain models are comparable line-by-line with the single domain model of the corresponding size (we observe similar results across all datasets). This suggests that, despite seeing 3 significantly different taxonomies with 48 aspect category labels during training, the multi-domain models are nevertheless able to make predictions grounded on the specific input prompt during inference.

\subsubsection{Post-processing non-extractive model predictions}
\label{subsubsec:attributingnonextractive}

Since model predictions containing a ``hallucinated'' substring that does not appear in the input text will be systematically scored as incorrect, thereby making inter-model comparisons difficult and causing downstream errors in OpinionTool\ visualization interfaces, after obtaining Table~\ref{tab:resultsfailuremode} we developed a model-independent similarity-based method to correct these predictions and standardize subsequent evaluations.

Given an input such as \textit{``I love New Yrok Citee --- it is the best!''}, here with spelling mistakes, and supposing that a model predicts \textit{``New York City''} as a target or opinion expression, our pipeline flags that the raw prediction cannot be located within the input text. In such cases, we create a list of all possible valid spans in the input: \texttt{[``I'', ``I love'', ..., ``New'', ``New Yrok'',``New Yrok Citee'', ...]}. Then, using the Levenshtein distance\footnote{\url{https://github.com/rapidfuzz/RapidFuzz}}, we compare the model prediction against all of these valid spans and select the span with the highest similarity score. Based on a test suite of dozens of common error types encountered during experiments, we found that this approach successfully corrects almost all non-extractive model predictions.

\subsection{Task performance}

We present our results for all models and dataset combinations in Table~\ref{tab:resultsinference}. We focus on the results based on the strict evaluation criterion, however all conclusions discussed in this section also apply to the relaxed evaluation which produces an improvement of 10 points on average for all models. Example model predictions are shown in Appendix~\ref{subsec:modelpredictionexamples}, and the score breakdown per language is presented in Table~\ref{tab:results-language-breakdown}.

\begin{table}[ht!]
\small
\centering
\setlength{\tabcolsep}{4pt}
\begin{tabular}{@{}lcccccccc}
\toprule
&  & \textbf{PS} & \textbf{HR} & \textbf{CR} \\
\midrule
GPT-4o-m. & NA & 4.88/10.89 & 6.45/19.62 & 12.11/20.46  \\
\hline
Base & 0.5B & 0.15/0.35 & 0.00/0.35 & 0.52/3.32  \\
& 1.5B & 0.79/2.24 & 0.47/3.87 & 1.76/5.08    \\
& 3B & 1.75/6.44 & 2.17/10.91 & 5.11/13.39  \\
\hline
Single d. & 0.5B & 23.77/35.02 & 12.31/18.46 & 30.34/37.69 \\
& 1.5B & 27.30/39.03 & 15.35/23.81 & 33.08/41.40 \\
& 3B & 31.47/43.83 & 21.99/30.68 & 39.38/49.30 \\
\hline
Multi-d. & 0.5B & 22.63/33.64 & 13.14/19.97 & 31.03/40.92 \\
& 1.5B & 27.51/39.36 & 17.66/26.16 & 35.40/45.62 \\
& 3B & 30.61/42.46 & 19.29/26.78 & 40.52/49.87 \\
\bottomrule
\end{tabular}
\caption{\textbf{Strict/Relaxed} F1 score. Columns represent the 3 datasets. Base models are not fine-tuned; other models are trained with English prompts. Single domain results are obtained from 3 distinct models; for example, the single domain model trained only on PS domain and evaluated on that same PS domain obtains F1 score of 23.77. Multi-domain results across the 3 dataset columns are obtained from the same multi-domain model.}
\label{tab:resultsinference}
\end{table}

Focusing first on the base model performance, we note that task F1 strict evaluations are close to 0 across all test datasets. Examining the baseline with GPT-4o mini we note that, while scores are relatively better than the largest of our base models, for our purposes the absolute performance is still unsatisfactory which provides a strong argument for the use of custom fine-tuned models for our internal ABSA use case.

Moving to the fine-tuned models, we observe several noteworthy results. All models fine-tuned on single domains show marked performance increases over both their corresponding base model, and over GPT-4o mini, with even our smallest 0.5B trained model showing a 15 point improvement on average. Within each dataset we also observe a clear monotonic trend with increasing model size. 

Comparing single and multi-domain models, we see that for all model sizes the multi-domain model achieves performances essentially identical to the specialized models, which allows us to support the practical recommendation of using the multi-domain 3B model in production for OpinionTool. These results also align with our related experiments on public datasets described in Appendix~\ref{app:m-absa-experiments}.

We note that there is a generally consistent ``difficulty'' pattern for the 3 datasets, with CR scoring consistently higher than PS, and HR. As outlined in Table~\ref{tab:datasetstable} the HR dataset displays a more formal writing style, and refers essentially to a highly specific working context and company vocabulary, while the CR domain is closer to generic datasets that probably appear in models' pre-training, and displays simpler and more direct opinion expressions.

Lastly, customizing the language of prompts to match the language of the samples resulted in moderate performance improvements. The highest gains (up to 3 points for HR) were observed when using a combination of English and French prompts, compared to using English-only instruction data. Detailed results are presented in Table~\ref{tab:prompt-language-results}.

\section{Conclusion}
\label{sec:conclusion}

In this work we have explored a variant of the novel ABSA quadruple task, focusing on complex \textit{opinion expressions}, and originating in the specific daily use cases of analysts in our company.
From our experiments with multilingual datasets, spanning application domains that are uncommon in public datasets, we have developed and evaluated a generic pipeline that can be used to reliably integrate a single, multi-domain 0.5-3B parameter LLM into real-world sentiment analysis systems, while achieving significant performance improvements over closed-source GPT-4o mini. Our results show that developing a single ABSA model is operationally feasible and effective, significantly reducing the effort required for model maintenance and deployment.

\section*{Limitations}

We focused on three internal datasets for quadruple extraction and one public dataset for triple extraction that was closest to our intended use case, however adapting public datasets by annotating complex opinion expressions for the quadruple extraction task could provide further evidence for the ability of small LLMs to perform well in quadruple ABSA tasks across multiple domains and multiple languages.

We noted that the datasets used during training are imbalanced both in distribution of their sentiment labels and their aspect category labels. Given that under-sampling real data would result in a dataset that is too small for effective supervised fine-tuning of small LLMs, we are considering developing data augmentation approaches and generating synthetic data for the under-represented class combinations in our datasets.

\bibliography{custom}

\clearpage
\appendix

\begin{table*}[ht!]
\section{Appendix}
\label{sec:appendix}
\subsection{Dataset statistics}
\label{subsec:datasetstats}
\setlength{\tabcolsep}{5pt}  
\small
\centering
\begin{tabular}{llrrrrrrr|rr|r}
\toprule
& & \multicolumn{7}{c|}{Languages} & \multicolumn{2}{c|}{Targets} & \\
\cline{3-12}
 &  & fr & en & ro & es & ar & pl & nl & Implicit & Explicit & Avg. \# quads \\
Dataset & Split &  &  &  &  &  &  &  &  &  &  \\
\midrule
\multirow[t]{3}{*}{PS} & train & 2,047 & 286 & 1,469 & 1,488 & 1,185 & 864 & 594 & 6,278 & 10,933 & 2.17 \\
 & validation & 693 & 96 & 490 & 497 & 398 & 289 & 198 & 2,076 & 3,629 & 2.14 \\
 & test & 688 & 96 & 490 & 497 & 398 & 289 & 198 & 2,079 & 3,720 & 2.18 \\
\cline{1-12}
\multirow[t]{3}{*}{HR} & train & 1,474 & 618 & - & - & - & - & - & 1,303 & 4,107 & 2.59 \\
 & validation & 491 & 209 & - & - & - & - & - & 465 & 1,589 & 2.93 \\
 & test & 495 & 212 & - & - & - & - & - & 450 & 1,411 & 2.63 \\
\cline{1-12}
\multirow[t]{3}{*}{CR} & train & 2,285 & - & - & - & - & - & - & 1,932 & 3,291 & 2.29 \\
 & validation & 761 & - & - & - & - & - & - & 593 & 1,149 & 2.29 \\
 & test & 763 & - & - & - & - & - & - & 572 & 1,013 & 2.08 \\
\bottomrule
\end{tabular}
\caption{Dataset statistics by split and by domain. We show the number of samples for each language and for each of the 3 domains. The \textit{Targets} columns counts the number of implicit (\textit{NULL} as referred to in the main text) and explicit opinion targets appearing in all samples. Finally we show the average number of quads per sample in the dataset.}
\label{tab:datasetstats1}
\end{table*}

\begin{table*}[b!]
\setlength{\tabcolsep}{5pt}  
\centering
\begin{tabular}{llrrrrrr}
\toprule
Dataset & Split &  0 & 1 & 2 & 3 & 4 & 5+ \\
\midrule
\multirow[t]{3}{*}{PS} & train & 173 & 3,193 & 2,173 & 1,207 & 569 & 618 \\
 & validation & 60 & 1,081 & 739 & 375 & 203 & 203 \\
 & test & 53 & 1,065 & 725 & 387 & 197 & 229 \\
\cline{1-8}
\multirow[t]{3}{*}{HR} & train & 35 & 750 & 535 & 340 & 176 & 256 \\
 & validation & 13 & 215 & 164 & 133 & 72 & 103 \\
 & test & 13 & 252 & 164 & 127 & 56 & 95 \\
\cline{1-8}
\multirow[t]{3}{*}{CR} & train & 51 & 684 & 761 & 448 & 180 & 161 \\
 & validation & 16 & 230 & 251 & 152 & 61 & 51 \\
 & test & 17 & 281 & 254 & 127 & 45 & 39 \\
\bottomrule
\end{tabular}
\caption{Detailed frequency distribution of the number of quads per sample; we note that most samples typically contain 1 or 2 quads.}
\label{tab:datasetstats2}
\end{table*}

\clearpage
\subsection{Prompt Modeling}
\label{subsec:sampleprompt}

For all prompts used during the development of our system, we standardize all quadruples for a given sample so that the opinion expressions that are to be predicted appear in the same order of appearance as in the input text. This can help during the post-processing of LLM outputs to attribute model predictions to spans in the input text, especially for long texts, if needed in visualization interfaces for example.

We show here the exact single domain prompt used for the Products and Services (PS) domain. The \textit{example document} is created manually based on familiarity with the contents of the domain, and is a representative example---both in terms of number of quadruples and in terms of style--- of the dataset contents.

The single domain prompts for the other datasets used in all experiments are exactly identical in structure to this one, but with 2 content changes: firstly, we replace the content of the \textit{allowed aspect categories} list with the corresponding list for the domain being studied. Secondly, we create the corresponding \textit{example document} for the domain exactly as described above.

Finally, for multi-domain experiments, we reuse the exact 3 sets of single domain prompts; we merge these 3 instruction datasets of prompts into one large dataset, then shuffle the samples with a fixed seed, and store this resulting multi-domain instruction dataset for use in all subsequent experiments.

\begin{figure*}[ht!]
\centering
\begin{tcolorbox}[colback=white, colframe=black, title={PS domain prompt}]
Below is a DOCUMENT in which human beings may be expressing themselves about products or services.

Perform a full aspect-based sentiment analysis of the DOCUMENT.

Only use sentiment labels that appear in the below list of ALLOWED SENTIMENTS.

Only use aspect category labels that appear in the below list of ALLOWED ASPECT CATEGORIES. Read the description of each label carefully to decide which one to use.

Use the exact words and spelling found in the DOCUMENT without modification.

Follow the FORMATTING EXAMPLE below, and return your answer as a structured JSON object without deviation.

DOCUMENT

\{{\texttt{example\_text}}\}

ALLOWED SENTIMENTS

- negative

- positive

- neutral

- mixed

ALLOWED ASPECT CATEGORIES

- evolution (description: use this label if the customer is expressing an opinion about a change in the product)

- price (description: use this label if the customer is expressing an opinion about the price of a product or service)

- reliability (description: use this label if the customer is expressing an opinion about the reliability of a product or service)

- suitability\_to\_needs (description: use this label if the customer is expressing an opinion about how well the product or service meets their needs)

- usability (description: use this label if the customer is expressing an opinion about the usability of a product or service)

- aesthetics (description: use this label if the customer is expressing an opinion about the aesthetic or appearance of a product or service)

$<...>$

FORMATTING EXAMPLE

Example document: My new TV never breaks down, but I think that the app store is too expensive.

Example structured JSON object answer: 

\{"aspect\_based\_sentiment\_analysis": [\{"target": "TV", "aspect\_category": "reliability", "sentiment": "positive", "opinion\_expression": "My new TV never breaks down"\}, \{"target": "app store", "aspect\_category": "price", "sentiment": "negative", "opinion\_expression": "the app store is too expensive"\}]\}

RESPONSE
\end{tcolorbox}
\end{figure*}

\clearpage
\subsection{Fine-tuning hyperparameters}
\label{app:finetuninghyperparams}

We trained all models for 3 epochs with an AdamW optimizer, a learning rate of $2 \times 10^{-5}$, and with early stopping on the validation loss. Hyperparameters were selected based on previous experiments with the Qwen2.5 family; a detailed list of our training configuration is given in Table~\ref{tab:finetuninghyperparams}. All models were trained on A100 GPUs, using HuggingFace \texttt{transformers} \cite{wolf-etal-2020-transformers}.

\begin{table}[h!]
\setlength{\tabcolsep}{2pt}  
  \centering
  \begin{tabular}{rc}
    \hline
    \textbf{Parameter} & \textbf{Value} \\
    \hline
    learning\_rate      & \texttt{2.0e-05}            \\
    optim     & \texttt{paged\_adamw\_32bit}           \\
    lr\_scheduler\_type     & \texttt{cosine}           \\
    gradient\_accumulation\_steps     & \texttt{8}           \\
    gradient\_checkpointing & \texttt{True} \\
    warmup\_ratio & \texttt{0.05} \\
    weight\_decay & \texttt{0.1} \\
    per\_device\_train\_batch\_size & \texttt{16} \\
    bf16     & \texttt{True}           \\\hline
  \end{tabular}
  \caption{Hyperparameter settings used for all fine-tuning experiments.}
  \label{tab:finetuninghyperparams}
\end{table}

\subsection{Pydantic schema for GPT-4o mini requests}
\label{subsec:gptdataschema}

Note that in the code below the variables \texttt{ASPECT\_CATEGORIES} and \texttt{SENTIMENTS} are lists of strings representing the valid labels, from each sample's corresponding dataset, that are passed to this schema for each request to GPT-4o mini.

\begin{lstlisting}
class ComplexSentimentAnalysisTerm(BaseModel):

    target: str = Field(description="A specific word or short extract, taken without modification, from the context sentence which someone is expressing an opinion about. For example in the sentence \'I love my new iPhone, it is so great \', the target is iPhone. Use the specific word NULL if there is no explicit target, for example in the sentence \'Pretty good\', there is no explicit target.")
    
    aspect_category: Literal[tuple(ASPECT_CATEGORIES)] = Field(description="The best description of the category that the target belongs to. For example, if the person is talking about the price of a service, the aspect_category is: price")
    
    sentiment: Literal[tuple(SENTIMENTS)] = Field(description="The best description of the sentiment that the opinion towards the target belongs to. For example, if the person is satisfied with the service, the sentiment is: positive")
    
    opinion_expression: str = Field(description="A short extract, taken without modification, from the context sentence which explains or justifies the overall classification of the sentiment and aspect_category")

class AspectBasedSentimentAnalysis(BaseModel):

    aspect_based_sentiment_analysis: 
    
    List[ComplexSentimentAnalysisTerm] = Field(description="A structured list of individual topics or subjects in a sentence, which are then used for targeted sentiment analysis")
\end{lstlisting}

\clearpage

\begin{table*}[t]
\subsection{Model prediction examples} 
\label{subsec:modelpredictionexamples}
\small
    \centering
    \begin{tabular}{p{5cm}p{3cm}p{3cm}p{4cm}}
        \toprule
        \textbf{Text} & \textbf{Reference} & \textbf{Prediction} & \textbf{Comments} \\
        \midrule
        \rowcolor{green!70!yellow!40} \multirow{2}{*}{\parbox{0.0001cm}}{User friendly and enables u to do any thing u want any time a day} & NULL, User friendly, \texttt{Usability}, \textit{positive} & NULL, User friendly, \texttt{Usability}, \textit{positive} & \multirow{2}{*}{\parbox{0.0001cm}}{Perfect model prediction, despite implicit target (here the user is discussing an app) and several spelling mistakes} \\
        \rowcolor{green!70!yellow!40} & NULL, enables u to do any thing u want any time a day, \texttt{Suitability\_to\_needs}, \textit{positive} & NULL, enables u to do any thing u want any time a day, \texttt{Suitability\_to\_needs}, \textit{positive} & \\
        \rowcolor{green!70!yellow!40} & & & \\ 
        \rowcolor{green!70!yellow!40} \multirow{3}{*}{\parbox{0.0001cm}}{It is making things easier for me but Sometimes it doesn't work Getting better :smiling: :thumbs-up:} & NULL, It is making things easier for me, \texttt{Suitability\_to\_needs}, \textit{positive} & NULL, It is making things easier for me, \texttt{Suitability\_to\_needs}, \textit{positive} & \multirow{3}{*}{\parbox{0.0001cm}}{Perfect model prediction on a complex example with informal style, emojis, and quickly changing sentiments} \\
        \rowcolor{green!70!yellow!40} & NULL, Sometimes it doesn't work, \texttt{Reliability}, \textit{negative} & NULL, Sometimes it doesn't work, \texttt{Reliability}, \textit{negative} & \\
        \rowcolor{green!70!yellow!40} & NULL, Getting better, \texttt{Evolution}, \textit{positive} & NULL, Getting better, \texttt{Evolution}, \textit{positive} & \\
        & & & \\ 
        The app after latest update is crashing everytime i open it ! & latest update, The app after latest update is crashing everytime i open it, \texttt{Evolution}, \textit{negative} & app, The app after latest update is crashing everytime i open it, \texttt{Reliability}, \textit{negative} & Subtle difference between reference and model choice of target (the update vs the app itself), and consequent aspect category classification \\
        & & & \\ 
        \multirow{2}{*}{\parbox{0.0001cm}}{While paying with visa card the app loading without any response. There is no need from it.} & app, the app loading without any response, \texttt{Reliability}, \textit{negative} & NULL, While paying with visa card the app loading without any response, \texttt{Reliability}, \textit{negative} & \multirow{2}{*}{\parbox{0.0001cm}}{Model error on the opinion target, and generates an extra prediction (fails to capture the nuanced meaning of the word ``need'')} \\
        & & NULL, There is no need from it, \texttt{Suitability\_to\_needs}, \textit{negative} & \\
        & & & \\ 
        \multirow{2}{*}{\parbox{0.0001cm}}{faire plus de formation en interne developper le partage de connaissance} & formation en interne, faire plus de formation en interne, \texttt{Tutoring}, \textit{negative} & formation en interne, faire plus de formation en interne, \texttt{Tutoring}, \textit{neutral} & \multirow{2}{*}{\parbox{0.0001cm}}{French language example from HR domain (transl. ``do more internal training, develop knowledge sharing'') where disagreement occurs for sentiment fields only---annotator considers suggestions to be negative while model considers them neutral} \\
        & partage de connaissance, developper le partage de connaissance, \texttt{Collaboration}, \textit{negative} & partage de connaissance, developper le partage de connaissance, \texttt{Collaboration}, \textit{neutral} & \\
        
        \bottomrule
    \end{tabular}
    \caption{Example model predictions (3B multi-domain fine-tuned model) illustrating several representative samples across our datasets. The highlighted parts of the table show perfect predictions.}
    \label{tab:examplemodeloutputs}
\end{table*}

\clearpage

\subsection{Experiments With M-ABSA Benchmark}
\label{app:m-absa-experiments}

\begin{table}[h!]
\small
\centering
\setlength{\tabcolsep}{4pt}
\begin{tabular}{@{}lcccccccc}
\toprule
&  \textbf{Lang.} & \textbf{Coursera} & \textbf{Food} & \textbf{Phone} \\
\midrule
Single d. & ar & 59.06 & 50.20 & 57.53 \\
& en & 73.88 & 54.20 & 56.36 \\
& es & 60.61 & 49.72 & 54.00 \\
& fr & 69.46 & 55.89 & 55.90 \\
& nl & 73.81 & 51.48 & 55.36 \\
\hline
Multi-d. & ar & 61.70 & 56.47 & 55.75 \\
& en & 81.87 & 60.37 & 57.65 \\
& es & 62.58 & 57.95 & 52.96 \\
& fr & 75.90 & 59.83 & 57.55 \\
& nl & 76.25 & 54.39 & 53.20 \\
\bottomrule
\end{tabular}
\caption{Strict F1 scores by language for Qwen2.5 3B Instruct models finetuned on the public M-ABSA dataset. We show 3 separate single domain models finetuned on each the Coursera, Food, and Phone domains individually, and 1 multi-domain model finetuned on all domains combined then evaluated on each domain individually.}
\label{tab:mabsaresults}
\end{table}

As outlined briefly in the main text, M-ABSA \citep{wu2025mabsa} is a recent multilingual ABSA dataset ranging over 7 different domains and 21 languages. Given our focus on the novel task of extracting opinion expressions, corresponding to our internal industrial needs and our consequent annotation efforts on private datasets, we were unable to find a public dataset with the exact data format match to ours. Therefore we identified the M-ABSA dataset as a close approximation, with which to validate our methodology: it contains almost all of the languages that appear in our internal data (except Polish and Romanian), and it contains varied domains with specific vocabulary, sentence length distributions, and aspect categories. However, the dataset only contains annotations for the ABSA-triple subtask, namely the annotations of the target, sentiment, and aspect category triples that occur in each sample sentence.

We prepared the dataset for our training pipeline by performing the following preprocessing steps. We normalized all the sentiment annotations to be either: positive, negative, neutral (some raw samples had labels such as POS or NEG). We filtered a few  samples in rare cases where we could not match the annotated target substring within the original sample text during processing of the raw dataset. We took the initial 30 detailed aspect categories of the Coursera domain (listed in Appendix H of \citet{wu2025mabsa}) and merged them to only keep their corresponding 7 high-level aspect category labels, to align more closely with the statistics of our private datasets. The Food domain contains a total of 10 different aspects, the Coursera domain contains 7 aspects, and the Phone domain contains 24. We show in Table~\ref{tab:mabsastats} an overview of the statistics of our M-ABSA dataset obtained after these steps.

The remaining steps of the training protocol with the M-ABSA datasets are identical to our main supervised finetuning experiments described in the main text. We use an identical prompt layout (up to a change of the aspect category labels in the prompt), and train on samples containing only the 3 keys of target, sentiment, and aspect category, omitting any information about the opinion expression task. We use the same training arguments and hyperparameters from Appendix~\ref{app:finetuninghyperparams}.

To validate the approach of training on multiple domains and multiple languages, we first fine-tune 3 single domain models on each of the 3 separate domains (Coursera, Food, Phone). We then create a multi-domain model that is trained on all 3 domains combined. In Table~\ref{tab:mabsaresults} we show the Strict F1 scores for all of the 3B parameter single domain finetuned models, and the scores for the multi-domain model evaluated on each of the 3 individual domains. As found for our private industrial datasets, we note firstly that the multi-domain model almost always exceeds the performance of the corresponding single domain model, across all languages (with small exceptions being the Phone domain in Arabic and Spanish).

Regarding the multilingual performance, we find similar trends as with our private industrial datasets (summarized in Table~\ref{tab:results-language-breakdown}) where performance on English examples is consistently better than on other languages; in this case with the M-ABSA datasets, this could also be due to the automatic translation quality from the original English, or the general influence of large amounts of English data in the Qwen models' training mix.

We note that, for comparison purposes, in Table 31 of \citet{wu2025mabsa} the authors report their own zero-shot baselines using Qwen2.5 7B: with our finetuning of the much smaller model from the same family, we find that our performances exceed these baselines (although we note again that we have used a significantly reduced number of aspect categories).

Finally we note that, like all public datasets, it is possible that the datasets in M-ABSA have been included in the Qwen2.5 models' pretraining; this may provide a simple explanation for why performance on all ABSA tasks decreases when implemented with (unseen) private industrial datasets.

\begin{table*}[h!]
\begin{tabular}{llrrrrr|rr|r}
\toprule
 &  & fr & en & es & ar & nl & Implicit & Explicit & Average number of triples \\
Dataset & Split &  &  &  &  &  &  &  &  \\
\midrule
\multirow[t]{3}{*}{coursera} & train & 1271 & 1278 & 1277 & 1276 & 1256 & 2116 & 3823 & 0.93 \\
 & validation & 332 & 336 & 336 & 336 & 331 & 564 & 1267 & 1.10 \\
 & test & 563 & 566 & 566 & 566 & 553 & 1312 & 2377 & 1.31 \\
\cline{1-10}
\multirow[t]{3}{*}{food} & train & 1268 & 1278 & 1273 & 1274 & 1266 & 3490 & 1849 & 0.84 \\
 & validation & 310 & 312 & 311 & 311 & 311 & 1045 & 640 & 1.08 \\
 & test & 511 & 522 & 522 & 521 & 521 & 2067 & 1622 & 1.42 \\
\cline{1-10}
\multirow[t]{3}{*}{phone} & train & 1053 & 1276 & 1274 & 1275 & 1265 & 0 & 13586 & 2.21 \\
 & validation & 258 & 307 & 307 & 306 & 302 & 0 & 3323 & 2.25 \\
 & test & 422 & 522 & 522 & 522 & 518 & 0 & 5689 & 2.27 \\
\bottomrule
\end{tabular}
\caption{Dataset statistics by split and by domain for the public M-ABSA dataset. We show the number of samples for each language and for each of the 3 domains after our preprocessing steps. Note that, as described in Section 4.3 of \citet{wu2025mabsa}, the Phone domain is not annotated with any implicit aspects.}
\label{tab:mabsastats}
\end{table*}

\clearpage

\begin{table*}[h!]
\centering
\setlength{\tabcolsep}{4.5pt}
\begin{tabular}{lllllllll}
\toprule
 &   & ar & en & es & fr & nl & pl & ro \\
model & dataset &  &  &  &  &  &  &  \\
\midrule
\multirow[t]{3}{*}{0.5B Multi} & PS & 0.2/0.32 & 0.47/0.51 & 0.22/0.34 & 0.34/0.43 & 0.24/0.33 & 0.14/0.27 & 0.17/0.28 \\
 & CR & - & - & - & 0.31/0.41 & - & - & - \\
 & HR & - & 0.11/0.19 & - & 0.14/0.2 & - & - & - \\
\cline{1-9}
0.5B Single & PS & 0.22/0.35 & 0.45/0.5 & 0.24/0.36 & 0.32/0.4 & 0.28/0.38 & 0.17/0.32 & 0.18/0.28 \\
\cline{1-9}
0.5B Single & CR & - & - & - & 0.3/0.38 & - & - & - \\
\cline{1-9}
0.5B Single & HR & - & 0.09/0.15 & - & 0.14/0.2 & - & - & - \\
\cline{1-9}
\multirow[t]{3}{*}{1.5B Multi} & PS & 0.27/0.4 & 0.5/0.59 & 0.28/0.4 & 0.39/0.47 & 0.28/0.4 & 0.19/0.36 & 0.19/0.31 \\
 & CR & - & - & - & 0.35/0.46 & - & - & - \\
 & HR & - & 0.15/0.24 & - & 0.19/0.27 & - & - & - \\
\cline{1-9}
1.5B Single & PS & 0.26/0.41 & 0.47/0.53 & 0.27/0.4 & 0.37/0.45 & 0.28/0.38 & 0.22/0.38 & 0.21/0.31 \\
\cline{1-9}
1.5B Single & CR & - & - & - & 0.33/0.41 & - & - & - \\
\cline{1-9}
1.5B Single & HR & - & 0.12/0.2 & - & 0.16/0.25 & - & - & - \\
\cline{1-9}
\multirow[t]{3}{*}{3B Multi} & PS & 0.29/0.44 & 0.54/0.62 & 0.31/0.43 & 0.4/0.48 & 0.33/0.44 & 0.22/0.39 & 0.25/0.35 \\
 & CR & - & - & - & 0.41/0.5 & - & - & - \\
 & HR & - & 0.16/0.24 & - & 0.2/0.28 & - & - & - \\
\cline{1-9}
3B Single & PS & 0.29/0.44 & 0.52/0.57 & 0.31/0.44 & 0.41/0.49 & 0.36/0.46 & 0.25/0.44 & 0.25/0.36 \\
\cline{1-9}
3B Single & CR & - & - & - & 0.39/0.49 & - & - & - \\
\cline{1-9}
3B Single & HR & - & 0.18/0.28 & - & 0.23/0.32 & - & - & - \\
\bottomrule
\end{tabular}
\caption{\textbf{Strict/Relaxed} F1 scores for all single and multi-domain models, analyzed by language of the test sample text. We show here results from the models trained using English instructions and prompts.}
\label{tab:results-language-breakdown}
\end{table*}

\clearpage

\begin{table*}[h!]
\centering
\setlength{\tabcolsep}{4.5pt}
\begin{tabular}{llllllllll}
\toprule
 &   & ar & en & es & fr & nl & pl & ro & overall\\
prompt & dataset &  &  &  &  &  &  &  & \\
\midrule
\multirow[t]{3}{*}{English} & PS & 0.29/0.44 & 0.54/0.62 & 0.31/0.43 & 0.40/0.48 & 0.33/0.44 & 0.22/0.39 & 0.25/0.35 & 0.31/0.42\\
 & CR & - & - & - & 0.41/0.50 & - & - & - & 0.41/0.50\\
 & HR & - & 0.16/0.24 & - & 0.20/0.28 & - & - & - & 0.19/0.27\\
\cline{1-10}
\multirow[t]{3}{*}{French} & PS & 0.30/0.42 & 0.47/0.59 & 0.33/0.44 & 0.42/0.50 & 0.32/0.44 & 0.23/0.40 & 0.23/0.35 & 0.30/0.43\\
 & CR & - & - & - & 0.40/0.49 & - & - & - & 0.40/0.49\\
 & HR & - & 0.15/0.26 & - & 0.23/0.32 & - & - & - & 0.21/0.30\\
\cline{1-10}
\multirow[t]{3}{*}{Mix} & PS & 0.30/0.44 & 0.54/0.64 & 0.33/0.44 & 0.42/0.50 & 0.33/0.44 & 0.24/0.42 & 0.24/0.37 & 0.32/0.44\\
 & CR & - & - & - & 0.41/0.51 & - & - & - & 0.41/0.51\\
 & HR & - & 0.18/0.30 & - & 0.24/0.32 & - & - & - & 0.22/0.31\\
\bottomrule
\end{tabular}
\caption{\textbf{Strict/Relaxed} F1 scores for Qwen2.5 3B multi-domain trained with English, French, and a mix of English and French prompts.}
\label{tab:prompt-language-results}
\end{table*}

\end{document}